# BLIND IMAGE DEBLURRING VIA REWEIGHTED GRAPH TOTAL VARIATION

*Yuanchao Bai*[⋆], *Gene Cheung*[†], *Xianming Liu*[$], *Wen Gao*[⋆]

[⋆]Peking University, Beijing, China, [$]Harbin Institute of Technology, Harbin, China
[†]National Institute of Informatics, Tokyo, Japan

## ABSTRACT

Blind image deblurring, *i.e.*, deblurring without knowledge of the blur kernel, is a highly ill-posed problem. The problem can be solved in two parts: i) estimate a blur kernel from the blurry image, and ii) given estimated blur kernel, de-convolve blurry input to restore the target image. In this paper, by interpreting an image patch as a signal on a weighted graph, we first argue that a skeleton image—a proxy that retains the strong gradients of the target but smooths out the details—can be used to accurately estimate the blur kernel and has a unique bi-modal edge weight distribution. We then design a reweighted graph total variation (RGTV) prior that can efficiently promote bi-modal edge weight distribution given a blurry patch. However, minimizing a blind image deblurring objective with RGTV results in a non-convex non-differentiable optimization problem. We propose a fast algorithm that solves for the skeleton image and the blur kernel alternately. Finally with the computed blur kernel, recent non-blind image deblurring algorithms can be applied to restore the target image. Experimental results show that our algorithm can robustly estimate the blur kernel with large kernel size, and the reconstructed sharp image is competitive against the state-of-the-art methods.

## 1. INTRODUCTION

Image blur is a common image degradation, which is usually modeled as

$$\mathbf{b} = \mathbf{x} \otimes \mathbf{k} + \mathbf{n}, \qquad (1)$$

where $\mathbf{b}$ is the blurry image, $\mathbf{x}$ is the latent sharp image, $\mathbf{k}$ is the blur kernel, $\mathbf{n}$ is the noise and $\otimes$ is the convolution operator. Image deblurring is to recover the latent sharp image $\mathbf{x}$ from the blurry image $\mathbf{b}$. We focus on the blind image deblurring problem [1], where both the latent image $\mathbf{x}$ and the blur kernel $\mathbf{k}$ are unknown and must be restored given only the blurry image $\mathbf{b}$. It is a highly ill-posed problem, since the feasible solution of the problem is not only unstable but also non-unique.

To overcome the ill-posedness, for blind image deblurring, it is important to design a prior that promotes image sharpness and penalizes blurriness. However, conventional gradient-based priors of natural images tend to fail [1], because they usually favor blurry images with mostly low frequencies in the Fourier domain. Recently, many sophisticated image priors are proposed to deal with this problem, for example, $L_0$ norm-based prior [2], low-rank prior [3] and dark channel prior [4]. Besides these priors, with the advance of *graph signal processing* (GSP) [5], graph-based priors have been designed for different image applications [6–8]. By modeling pixels as nodes with weighted edges that reflect inter-pixel similarities, images can be interpreted as signals on graphs. In this paper, we explore the relationship between graph and image blur, and propose a graph-based prior for blind image deblurring.

Specifically, instead of directly computing the natural image, we argue that a *skeleton image*—a piecewise smooth (PWS) proxy that retains the strong gradients of the target image but smooths out the details—is sufficient to estimate the blur kernel. We observe that, unlike blurry patches, the edge weights of a graph for the skeleton image patch have a unique bi-modal distribution. We thus propose a *reweighted graph Total Variation* (RGTV) prior to promote the desirable bi-modal distribution given a blurry patch. We juxtapose and analyze the advantages of RGTV against previous graph smoothness priors, such as graph TV (GTV) [9] and the graph Laplacian regularizer [7]. We propose a fast algorithm that solves for the skeleton image and the blur kernel alternately. Then with the estimated blur kernel $\mathbf{k}$, we de-convolve the blurry image using a non-blind deblurring method, like [2, 10, 11]. Experiments show that our algorithm achieves competitive or better results than many state-of-the-art methods.

The outline of the paper is as follows. We introduce the graph definition and observation in Sec. 2. RGTV prior and blind deblurring algorithm are proposed in Sec. 3 and Sec. 4. Experiments and conclusions are in Sec. 5 and Sec. 6.

## 2. GRAPH-BASED IMAGE PRIOR

### 2.1. Graph Definition

We first introduce definitions of a graph. A graph $\mathcal{G}(\mathcal{V}, \mathcal{E}, \mathbf{W})$ is a triplet consisting of a finite set of $\mathcal{V}$ of $N$ nodes (im-age pixels) and a finite set $\mathcal{E} \subset \mathcal{V} \times \mathcal{V}$ of $M$ edges. Each edge $(i,j) \in \mathcal{E}$ is undirected with a corresponding weight $w_{ij}$ which measures the similarity between nodes $i$ and $j$. Here we compute the weights using a Gaussian kernel [5]:

$$[\mathbf{W}]_{i,j} = w_{i,j} = \exp(-\frac{\|x_i - x_j\|^2}{\sigma^2}), \qquad (2)$$

where $\mathbf{W}$ is the graph weight matrix of size $N \times N$, $x_i$ and $x_j$ are the pixels $i$ and $j$ of the image $\mathbf{x}$, and $\sigma$ is a smoothing parameter. $0 \leq w_{ij} \leq 1$ and the larger $w_{ij}$ is, the more similar the nodes $i$ and $j$ are to each other.

Given the graph matrix $\mathbf{W}$, a *combinatorial graph Laplacian matrix* $\mathbf{L}$ is a symmetric matrix defined as:

$$\mathbf{L} \triangleq \mathrm{diag}(\mathbf{W1}) - \mathbf{W} \qquad (3)$$

where $\mathbf{1}$ is a vector of all 1s. $\mathrm{diag}(\cdot)$ is an operator constructing a square diagonal matrix with the elements of input vector on the main diagonal.

### 2.2. Observation of Bi-modal Distribution

As directly computing a target natural image without blur kernel is very difficult, we compute a *skeleton image*—a piecewise smooth (PWS) version of the target image as a proxy. The skeleton image retains the strong gradient of a natural image but smooths out the minor details, which is similar to a structure extracted image [12] or an edge-aware smoothed image [13]. An illustrative example is shown in Fig. 1. Both of the target natural image and its skeleton image are sharper than the blurry image in the middle.

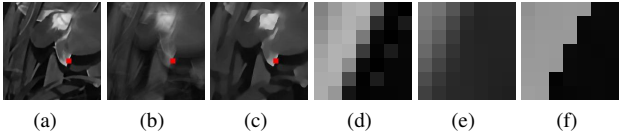

(a)　(b)　(c)　(d)　(e)　(f)

**Fig. 1**: Illustrations of different kinds of images. (a) a true natural image. (b) a blurry image. (c) a skeleton image. (d), (e) and (f) are patches in red squares of (a), (b) and (c), respectively.

Beyond visual differences, we seek also statistical descriptions of these images so that we can differentiate among them in a more mathematical rigorous manner. Specifically, we examine the edge weight distribution of a graph for each of three representative patches, where edge weight $w_{i,j}$ is computed using (2). Fig. 2 shows the edge weight distributions (histograms) of the representative patches in Fig. 1d–1f, where $d = |x_i - x_j|$ is the $x$-coordinate. The fraction of weights is the $y$-coordinate. We make the following key observation from the histograms:

Both the target natural patch and its skeleton version have bi-modal distribution of edge weights, while the blurred patch does not, due to low-pass filtering during the blur process. Besides, the bi-modal distribution of the skeleton patch is more desirable without the effects of details. Based on this observation, we design a signal prior to *promote* a bi-modal distribution of edge weights to reconstruct a skeleton proxy of target natural patch given an observed blurry patch. This is the focus of the next section.

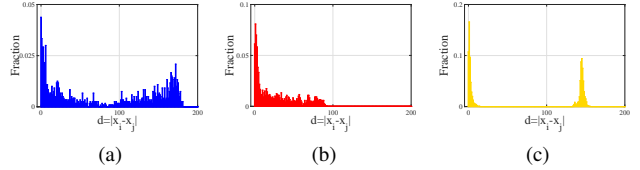

(a)　(b)　(c)

**Fig. 2**: Graph weight distribution properties around edges. (a) a true natural patch. (b) a blurry patch. (c) a skeleton patch.

## 3. GRAPH-BASED IMAGE REGULARIZATION

### 3.1. Reweighted Graph Total Variation Prior

We propose a *reweighted graph total variation* (RGTV) prior to promote the aforementioned desirable bi-modal edge weight distribution in a target pixel patch. We first define the gradient operator of a graph signal $\mathbf{x}$. The gradient of node $i \in \mathcal{V}$ is defined as $\nabla_i \mathbf{x} \in \mathbb{R}^N$ and its $j$-th element is:

$$(\nabla_i \mathbf{x})_j \triangleq x_j - x_i, \qquad (4)$$

The conventional graph TV (GTV) [9] is defined as

$$\|\mathbf{x}\|_{GTV} = \sum_{i \in \mathcal{V}} \|\mathrm{diag}(\mathbf{W}_{i,\cdot})\nabla_i \mathbf{x}\|_1$$
$$= \sum_{i=1}^{N} \sum_{j=1}^{N} w_{i,j}|x_j - x_i| \qquad (5)$$

where $\mathbf{W}_{i,\cdot}$ is the $i$-th row of the adjacency matrix $\mathbf{W}$. GTV initializes $\mathbf{W}$ using for example (2) and keeps it fixed, and hence does not promote bi-modal distribution. As (5) is separable, we can analyze the behavior of GTV using a single pair $(i,j)$ separately like a two-node graph. With $d = |x_i - x_j|$ and fixed $w_{i,j}$, the regularizer for pair $(i,j)$ is $w_{i,j}d$, which is a linear function of $d$ with slope $w_{i,j}$. The curve of $w_{i,j}d$ has only one minimum at $d = 0$, as shown in Fig. 3a. Minimizing (5) only pushes $d$ towards 0.

Instead of using fixed $\mathbf{W}$, we extend the conventional graph TV to the RGTV, where the graph weights $\mathbf{W}(\mathbf{x})$ are also functions of $\mathbf{x}$,

$$\|\mathbf{x}\|_{RGTV} = \sum_{i \in \mathcal{V}} \|\mathrm{diag}(\mathbf{W}_{i,\cdot}(\mathbf{x}))\nabla_i \mathbf{x}\|_1$$
$$= \sum_{i=1}^{N} \sum_{j=1}^{N} w_{i,j}(x_i,x_j)|x_j - x_i|, \qquad (6)$$

where $\mathbf{W}_{i,\cdot}(\mathbf{x})$ is the $i$-th row of $\mathbf{W}(\mathbf{x})$ and $w_{i,j}(x_i,x_j)$ is the $(i,j)$ element of $\mathbf{W}(\mathbf{x})$. The extension makes a fundamental difference, because the regularizer for pair $(i, j)$ now becomes $w_{i,j}(x_i, x_j)|x_j - x_i| = \exp(-d^2/\sigma^2) \cdot d$. The curve of this regularizer has one maximum at $\sigma/\sqrt{2}$ and two minima at 0 and $+\infty$, as shown in Fig. 3a. Minimizing (6) reduces $d$ if $d$ is smaller than $\sigma/\sqrt{2}$ and magnifies $d$ if $d$ is larger than $\sigma/\sqrt{2}$. Thus RGTV regularizer can effectively promote the desirable bi-modal edge weight distribution.

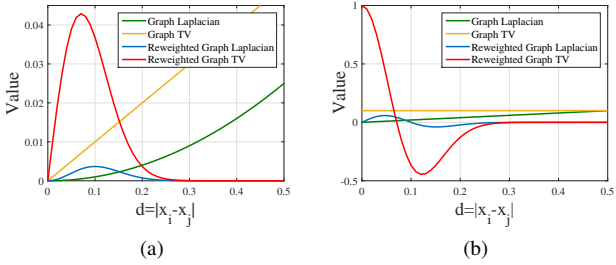

**Fig. 3**: Curves of regularizers and their corresponding first-derivatives for each $(i, j)$ pair. $d$ is normalized to $[0, 1]$. $w_{i,j} = 0.1$ for graph Laplacian and graph TV. $\sigma = 0.1$ for reweighted graph Laplacian and reweighted graph TV.

Using the aforementioned RGTV prior, we propose an optimization function for blind image deblurring in Sec. 4.

### 3.2. Comparisons with Graph Laplacian Prior

We analyze commonly used graph Laplacian regularizers. The graph Laplacian regularizer [7] is expressed as

$$\mathbf{x}^T \mathbf{L} \mathbf{x} = \sum_{i=1}^{N} \sum_{j=1}^{N} w_{i,j}(x_j - x_i)^2 \quad (7)$$

Like GTV, graph Laplacian initializes $\mathbf{L}$ and keeps it fixed, and hence does not promote the desirable bi-modal edge weight distribution. With $d = |x_i - x_j|$ and fixed $w_{i,j}$, the prior for each node pair $(i, j)$ is $w_{i,j} d^2$, which is a quadratic function of $d$ with coefficient $w_{i,j}$. The curve of $w_{i,j} d^2$ has only one minimum at $d = 0$, as shown in Fig. 3a. Minimizing (7) only pushes $d$ to 0.

We extend the conventional graph Laplacian to the reweighted graph Laplacian. Similar to RGTV, we define the reweighted graph Laplacian as

$$\mathbf{x}^T \mathbf{L}(\mathbf{x}) \mathbf{x} = \sum_{i=1}^{N} \sum_{j=1}^{N} w_{i,j}(x_i, x_j) \cdot (x_j - x_i)^2, \quad (8)$$

where Laplacian matrix $\mathbf{L}(\mathbf{x})$ is a function of $\mathbf{x}$ and $w_{i,j}(x_i, x_j)$ is the same as the definition in (6). Then, the regularizer for pair $(i, j)$ becomes $w_{i,j}(x_i, x_j)(x_j - x_i)^2 = \exp(-d^2/\sigma^2) \cdot d^2$. The curve has one maximum at $d = \sigma$ and two minima at 0 and $+\infty$, as shown in Fig. 3a. Though reweighted graph Laplacian can also promote the desirable bi-modal edge weight distribution based on Fig. 3a, it has one significant drawback. Taking the first derivative of its function results in $\exp(-d^2/\sigma^2) \cdot 2d(1 - d^2/\sigma^2)$, as shown in Fig. 3b. $\lim_{d \to 0} \exp(-d^2/\sigma^2) \cdot 2d(1 - d^2/\sigma^2) = 0$, which means that the convergence of minimizing (8) is very slow when $d$ is close to 0 in practice.

Different from reweighted graph Laplacian, the first derivative of the cost function of RGTV is $\exp(-d^2/\sigma^2) \cdot (1 - 2d^2/\sigma^2)$, as shown in Fig. 3b. $\lim_{d \to 0} \exp(-d^2/\sigma^2) \cdot (1 - 2d^2/\sigma^2) = 1$, which means that the RGTV can promote a bi-modal edge weight distribution and ensure a good convergence speed of minimizing (6) when $d$ is close to 0.

## 4. BLIND IMAGE DEBLURRING ALGORITHM

Using the blur image model in (1), we pose the blind image deblurring problem as an optimization as follows using our proposed RGTV prior:

$$\hat{\mathbf{x}}, \hat{\mathbf{k}} = \arg\min_{\mathbf{x}, \mathbf{k}} \frac{1}{2} \|\mathbf{x} \otimes \mathbf{k} - \mathbf{b}\|_2^2 + \lambda \|\mathbf{x}\|_{RGTV} + \mu \|\mathbf{k}\|_2^2 \quad (9)$$

where the first term is the data fidelity term, and the remaining two terms are regularization terms for variables $\mathbf{x}$ and $\mathbf{k}$, respectively. $\lambda$ and $\mu$ are two corresponding parameters.

The optimization (9) is non-convex and non-differentiable, which is challenging to solve. Here we apply an alternating scheme to solve the minimizers $\hat{\mathbf{x}}$ and $\hat{\mathbf{k}}$ iteratively, as sketched in Algorithm 1.

The minimizer $\hat{\mathbf{x}}$ is our PWS proxy—the skeleton image, in order to estimate a good blur kernel $\hat{\mathbf{k}}$. To restore the natural sharp image given estimated blur kernel $\hat{\mathbf{k}}$, we can use recent non-blind image deblurring algorithms to deblur the blurry image $\mathbf{b}$ such as [2, 10, 11].

### 4.1. Estimating Skeleton Image

Given $\hat{\mathbf{k}}$, optimization (9) to solve $\mathbf{x}$ becomes:

$$\hat{\mathbf{x}} = \arg\min_{\mathbf{x}} \frac{1}{2} \|\mathbf{x} \otimes \hat{\mathbf{k}} - \mathbf{b}\|_2^2 + \lambda \|\mathbf{x}\|_{RGTV} \quad (10)$$

RGTV is a non-differentiable prior, where the edge weights are functions of $\mathbf{x}$. We employ an alternating scheme again to separate $\mathbf{W}(\mathbf{x})$ from $\mathbf{x}$ optimization, i.e., we first optimize $\mathbf{x}$ with initialized $\mathbf{W}$ (of all ones), then we update $\mathbf{W}$ with $\mathbf{W}(\hat{\mathbf{x}})$ and optimize $\mathbf{x}$ again. Given $\mathbf{W}$ and $\hat{\mathbf{k}}$ to solve for $\mathbf{x}$, we solve the sub-problem with a primal-dual algorithm [14]. The alternating algorithm runs iteratively until convergence as the solution to (10).

### 4.2. Estimating Blur Kernel

To solve $\mathbf{k}$ given $\hat{\mathbf{x}}$, we make a slight modification and solve $\mathbf{k}$ in the gradient domain to avoid artifacts [15, 16]. The optimization (9) becomes:

$$\hat{\mathbf{k}} = \arg\min_{\mathbf{k}} \frac{1}{2} \|\nabla \hat{\mathbf{x}} \otimes \mathbf{k} - \nabla \mathbf{b}\|_2^2 + \mu \|\mathbf{k}\|_2^2 \quad (11)$$

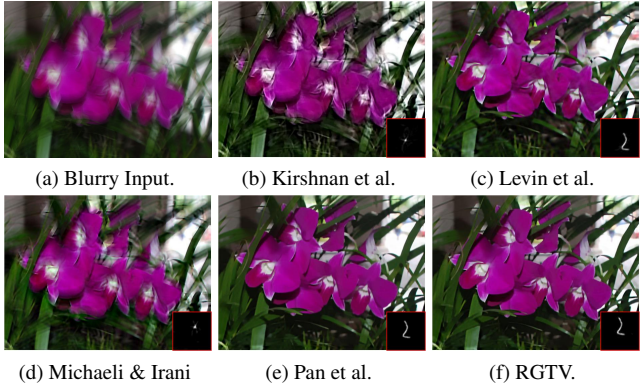

**Fig. 4**: *Flower*. Image size: $618 \times 464$, kernel size: $69 \times 69$.

where $\nabla$ is the gradient operator. (11) is a quadratic function and has a closed-form solver like deconvolution. We accelerate the solver via Fast Fourier Transforms [15]. After obtaining $\hat{\mathbf{k}}$, we threshold the negative elements to zeros and normalize $\hat{\mathbf{k}}$ to ensure $\sum_i \hat{k}_i = 1$.

---
**Algorithm 1** Solving (9)

---
**Input:** Blurry image $\mathbf{b}$ and kernel size $h \times h$.
1: Initialize $\hat{\mathbf{k}}$ with delta function or the result from coarser scale.
2: **while** not converge **do**
   Update $\hat{\mathbf{x}}$ by solving (10).
   Update $\hat{\mathbf{k}}$ by solving (11).
   $\lambda \leftarrow \lambda/1.1$.
   **endwhile**
**Output:** Estimated blur kernel $\hat{\mathbf{k}}$ and skeleton image $\hat{\mathbf{x}}$.

---

### 4.3. Coarse-to-Fine Strategy

We apply a coarse-to-fine strategy [17] to solve (9), in order to make the solver robust for large blur kernels. In the coarse-to-fine strategy, we construct an image pyramid by down-sampling the blurry image and do blind image deblurring scale by scale. In each scale, we estimate $\hat{\mathbf{k}}$ and $\hat{\mathbf{x}}$, and then we up-sample $\hat{\mathbf{k}}$ as the initial value for the finer scale.

## 5. EXPERIMENTS

In this section, we evaluate the performance of the proposed algorithm and compare it against four recent blind image deblurring problems, *i.e.*, Kirshnan et al. [18], Levin et al. [19], Michaeli & Irani [20], Pan et al. [4]. The experiments are applied on the challenging motion blurred natural images with large blur kernels. The codes of competing algorithms are offered by their authors online. All the algorithms are run on the Matlab 2015a.

For the proposed algorithm, the down-sampling factor for coarse-to-fine strategy is set to $\log_2 3$. We construct a four-neighbor adjacency graph on the image as a trade-off between performance and complexity. In (2), $\sigma = 0.1$. In (9), $\lambda$ is initialized to $0.01$ and $\mu$ is set to $0.05$. In blind image deblurring, the *kernel size* is unknown and is an important parameter. For fair comparisons, we set the same *kernel size* for all the algorithms to estimate the blur kernel in each case. Then, we use the same non-blind image deblurring algorithm in [2] to reconstruct sharp images with estimated blur kernels.

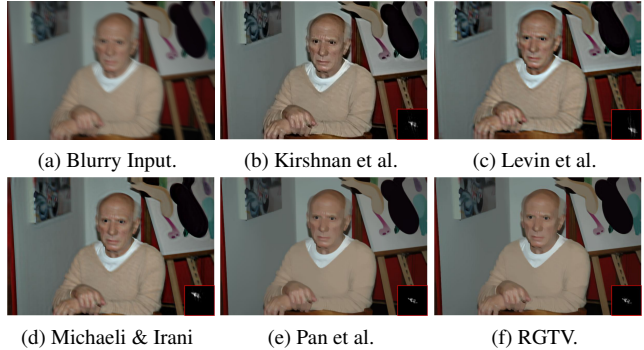

**Fig. 5**: *Picasso*. Image size: $800 \times 532$, kernel size: $69 \times 69$.

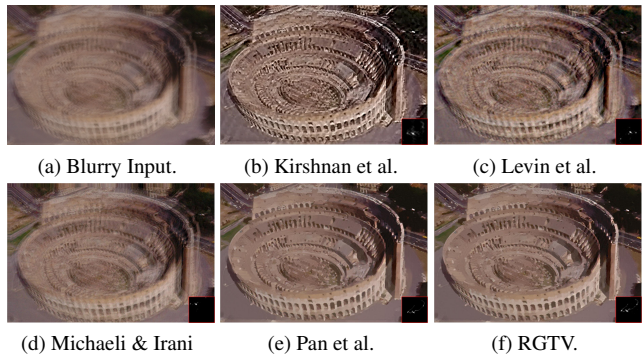

**Fig. 6**: *Roma*. Image size: $1229 \times 825$, kernel size: $95 \times 95$.

Fig. 4 – 6 demonstrate the estimated blur kernels and the corresponding deblurred images of all the algorithms. The proposed algorithm can estimate the blur kernels robustly and the visual quality of our reconstructed sharp images is competitive or better than that of the results of other algorithms. The images are better viewed in full size on computer screen.

## 6. CONCLUSION

RGTV is an effective prior to promote image sharpness and penalize blurriness. Based on the comparisons, it is better than other graph smoothness priors, *i.e.*, GTV and graph Laplacian, in solving blind image deblurring problem. In this paper, we design a fast algorithm with RGTV. Experimental results demonstrate that our algorithm can deal with large blur kernels and the reconstructed results are competitive against the state-of-the-art methods.

# 7. REFERENCES

[1] A. Levin, Y. Weiss, F. Durand, and W. T. Freeman., "Understanding blind deconvolution algorithms," *IEEE Transactions on Pattern Analysis and Machine Intelligence*, vol. 33, no. 12, pp. 2354–2367, Dec 2011.

[2] J. Pan, Z. Hu, Z. Su, and M. H. Yang, "L0-regularized intensity and gradient prior for deblurring text images and beyond," *IEEE Transactions on Pattern Analysis and Machine Intelligence*, vol. PP, no. 99, pp. 1–1, 2016.

[3] W. Ren, X. Cao, J. Pan, X. Guo, W. Zuo, and M. H. Yang, "Image deblurring via enhanced low-rank prior," *IEEE Transactions on Image Processing*, vol. 25, no. 7, pp. 3426–3437, 2016.

[4] Jinshan Pan, Deqing Sun, Hanspeter Pfister, and Ming-Hsuan Yang, "Blind image deblurring using dark channel prior," in *Proceedings of IEEE Conference on Computer Vision and Pattern Recognition*, June 2016.

[5] D. I. Shuman, S. K. Narang, P. Frossard, A. Ortega, and P. Vandergheynst, "The emerging field of signal processing on graphs: Extending high-dimensional data analysis to networks and other irregular domains," in *IEEE Signal Processing Magazine*, May 2013, vol. 30, no.3, pp. 83–98.

[6] W. Hu, G. Cheung, and M. Kazui, "Graph-based dequantization of block-compressed piecewise smooth images," *IEEE Signal Processing Letters*, vol. 23, no. 2, pp. 242–246, Feb 2016.

[7] J. Pang and G. Cheung, "Graph Laplacian regularization for inverse imaging: Analysis in the continuous domain," in *IEEE Transactions on Image Processing*, April 2017, vol. 26, no.4, pp. 1770–1785.

[8] X. Liu, G. Cheung, X. Wu, and D. Zhao, "Random walk graph laplacian-based smoothness prior for soft decoding of jpeg images," *IEEE Transactions on Image Processing*, vol. 26, no. 2, pp. 509–524, Feb 2017.

[9] P. Berger, G. Hannak, and G. Matz, "Graph signal recovery via primal-dual algorithms for total variation minimization," *IEEE Journal of Selected Topics in Signal Processing*, vol. 11, no. 6, pp. 842–855, Sept 2017.

[10] Dilip Krishnan and Rob Fergus, "Fast image deconvolution using hyper-laplacian priors," in *Proceedings of Neural Information Processing Systems*, 2009, pp. 1033–1041.

[11] Daniel Zoran and Yair Weiss, "From learning models of natural image patches to whole image restoration," in *Proceedings of IEEE International Conference on Computer Vision*, 2011, pp. 479–486.

[12] Li Xu, Qiong Yan, Yang Xia, and Jiaya Jia, "Structure extraction from texture via relative total variation," *ACM Trans. Graph.*, vol. 31, no. 6, pp. 1–10, 2012.

[13] Li Xu, Cewu Lu, Yi Xu, and Jiaya Jia, "Image smoothing via l0 gradient minimization," *ACM Trans. Graph.*, vol. 30, no. 6, pp. 1–12, 2011.

[14] Antonin Chambolle and Thomas Pock, "A first-order primal-dual algorithm for convex problems with applications to imaging," *J. Math. Imaging Vis.*, vol. 40, no. 1, pp. 120–145, 2011.

[15] Sunghyun Cho and Seungyong Lee, "Fast motion deblurring," *ACM Transactions on Graphics*, vol. 28, no. 5, pp. article no. 145, 2009.

[16] Li Xu and Jiaya Jia, "Two-phase kernel estimation for robust motion deblurring," in *Proceedings of European Conference on Computer Vision*, Berlin, Heidelberg, 2010, pp. 157–170.

[17] R. Fergus, B. Singh, A. Hertzmann, S. T. Roweis, and W.T. Freeman, "Removing camera shake from a single photograph," *ACM Transactions on Graphics*, vol. 25, pp. 787–794, 2006.

[18] D. Krishnan, T. Tay, and R. Fergus, "Blind deconvolution using a normalized sparsity measure," in *Proceedings of IEEE Conference on Computer Vision and Pattern Recognition*, Washington, DC, USA, 2011, pp. 233–240.

[19] A. Levin, Y. Weiss, F. Durand, and W. T. Freeman, "Efficient marginal likelihood optimization in blind deconvolution," in *Proceedings of IEEE Conference on Computer Vision and Pattern Recognition*, Washington, DC, USA, 2011, pp. 2657–2664.

[20] Tomer Michaeli and Michal Irani, "Blind deblurring using internal patch recurrence," in *Proceedings of European Conference on Computer Vision*, Cham, 2014, pp. 783–798.